# Neural Text Generation from Structured Data with Application to the Biography Domain

**Rémi Lebret**[*]
EPFL, Switzerland

**David Grangier**
Facebook AI Research

**Michael Auli**
Facebook AI Research

## Abstract

This paper introduces a neural model for concept-to-text generation that scales to large, rich domains. It generates biographical sentences from fact tables on a new dataset of biographies from Wikipedia. This set is an order of magnitude larger than existing resources with over 700k samples and a 400k vocabulary. Our model builds on conditional neural language models for text generation. To deal with the large vocabulary, we extend these models to mix a fixed vocabulary with *copy actions* that transfer sample-specific words from the input database to the generated output sentence. To deal with structured data, we allow the model to embed words differently depending on the data fields in which they occur. Our neural model significantly outperforms a Templated Kneser-Ney language model by nearly 15 BLEU.

## 1 Introduction

Concept-to-text generation renders structured records into natural language (Reiter et al., 2000). A typical application is to generate a weather forecast based on a set of structured meteorological measurements. In contrast to previous work, we scale to the large and very diverse problem of generating biographies based on Wikipedia infoboxes. An infobox is a fact table describing a person, similar to a person subgraph in a knowledge base (Bollacker et al., 2008; Ferrucci, 2012). Similar generation applications include the generation of product descriptions based on a catalog of millions of items with dozens of attributes each.

Previous work experimented with datasets that contain only a few tens of thousands of records such as WEATHERGOV or the ROBOCUP dataset, while our dataset contains over 700k biographies from Wikipedia. Furthermore, these datasets have a limited vocabulary of only about 350 words each, compared to over 400k words in our dataset.

To tackle this problem we introduce a statistical generation model conditioned on a Wikipedia infobox. We focus on the generation of the first sentence of a biography which requires the model to select among a large number of possible fields to generate an adequate output. Such diversity makes it difficult for classical count-based models to estimate probabilities of rare events due to data sparsity. We address this issue by parameterizing words and fields as embeddings, along with a neural language model operating on them (Bengio et al., 2003). This factorization allows us to scale to a larger number of words and fields than Liang et al. (2009), or Kim and Mooney (2010) where the number of parameters grows as the product of the number of words and fields.

Moreover, our approach does not restrict the relations between the field contents and the generated text. This contrasts with less flexible strategies that assume the generation to follow either a hybrid alignment tree (Kim and Mooney, 2010), a probabilistic context-free grammar (Konstas and Lapata, 2013), or a tree adjoining grammar (Gyawali and Gardent, 2014).

Our model exploits structured data both globally and locally. Global conditioning summarizes all information about a personality to understand high-level themes such as that the biography is about a scientist or an artist, while as local conditioning describes the previously generated tokens in terms of the their relationship to the infobox. We analyze the effectiveness of each and demonstrate their complementarity.

## 2 Related Work

Traditionally, generation systems relied on rules and hand-crafted specifications (Dale et al., 2003; Reiter et al., 2005; Green, 2006; Galanis and Androut-sopoulos, 2007; Turner et al., 2010). Generation is divided into modular, yet highly interdependent, decisions: (1) *content planning* defines which parts of the input fields or meaning representations should be selected; (2) *sentence planning* determines which selected fields are to be dealt with in each output sentence; and (3) *surface realization* generates those sentences.

---

[*] Rémi performed this work while interning at Facebook.

Data-driven approaches have been proposed to automatically learn the individual modules. One approach first aligns records and sentences and then learns a content selection model (Duboue and McKeown, 2002; Barzilay and Lapata, 2005). Hierarchical hidden semi-Markov generative models have also been used to first determine which facts to discuss and then to generate words from the predicates and arguments of the chosen facts (Liang et al., 2009). Sentence planning has been formulated as a supervised set partitioning problem over facts where each partition corresponds to a sentence (Barzilay and Lapata, 2006). End-to-end approaches have combined sentence planning and surface realization by using explicitly aligned sentence/meaning pairs as training data (Ratnaparkhi, 2002; Wong and Mooney, 2007; Belz, 2008; Lu and Ng, 2011). More recently, content selection and surface realization have been combined (Angeli et al., 2010; Kim and Mooney, 2010; Konstas and Lapata, 2013).

At the intersection of rule-based and statistical methods, hybrid systems aim at leveraging human contributed rules and corpus statistics (Langkilde and Knight, 1998; Soricut and Marcu, 2006; Mairesse and Walker, 2011).

Our approach is inspired by the recent success of neural language models for image captioning (Kiros et al., 2014; Karpathy and Fei-Fei, 2015; Vinyals et al., 2015; Fang et al., 2015; Xu et al., 2015), machine translation (Devlin et al., 2014; Bahdanau et al., 2015; Luong et al., 2015), and modeling conversations and dialogues (Shang et al., 2015; Wen et al., 2015; Yao et al., 2015).

Our model is most similar to Mei et al. (2016) who use an encoder-decoder style neural network model to tackle the WEATHERGOV and ROBOCUP tasks. Their architecture relies on LSTM units and an attention mechanism which reduces scalability compared to our simpler design.

| Frederick Parker-Rhodes | |
|---|---|
| Born | 21 November 1914 Newington, Yorkshire |
| Died | 2 March 1987 (aged 72) |
| Residence | UK |
| Nationality | British |
| Fields | Mycology, Plant Pathology, Mathematics, Linguistics, Computer Science |
| Known for | Contributions to computational linguistics, combinatorial physics, bit-string physics, plant pathology, and mycology |
| Author abbrev. (botany) | Park.-Rhodes |

**Figure 1:** Wikipedia infobox of Frederick Parker-Rhodes. The introduction of his article reads: "Frederick Parker-Rhodes (21 March 1914 – 21 November 1987) was an English linguist, plant pathologist, computer scientist, mathematician, mystic, and mycologist.".

## 3 Language Modeling for Constrained Sentence generation

Conditional language models are a popular choice to generate sentences. We introduce a table-conditioned language model for constraining text generation to include elements from fact tables.

### 3.1 Language model

Given a sentence $s = w_1, \ldots, w_T$ with $T$ words from vocabulary $\mathcal{W}$, a language model estimates:

$$P(s) = \prod_{t=1}^{T} P(w_t | w_1, \ldots, w_{t-1}). \quad (1)$$

Let $c_t = w_{t-(n-1)}, \ldots, w_{t-1}$ be the sequence of $n-1$ context words preceding $w_t$. An $n$-gram language model makes an order $n$ Markov assumption,

$$P(s) \approx \prod_{t=1}^{T} P(w_t | c_t). \quad (2)$$

### 3.2 Language model conditioned on tables

A table is a set of field/value pairs, where values are sequences of words. We therefore propose language models that are conditioned on these pairs.

**Local conditioning** refers to the information from the table that is applied to the description of the words which have already generated, i.e. the previous words that constitute the context of the language

| **input text** $(c_t, z_{c_t})$ | | | | | | | | | |
|---|---|---|---|---|---|---|---|---|---|
| | John | Doe | ( | 18 | April | 1352 | ) | is | a |
| $c_t$ | 13944 | unk | 17 | 37 | 92 | 25 | 18 | 12 | 4 |
| $z_{c_t}$ | (name,1,2) | (name,2,1) (spouse,2,1) (children,2,1) | ∅ | (birthd.,1,3) | (birthd.,2,2) | (birthd.,3,1) | ∅ | ∅ | ∅ |

**Table** $(g_f, g_w)$

| name | John Doe |
|---|---|
| birthdate | 18 April 1352 |
| birthplace | Oxford UK |
| occupation | placeholder |
| spouse | Jane Doe |
| children | Johnnie Doe |

**output candidates** $(w \in \mathcal{W} \cup \mathcal{Q})$

| | the | ... | april | ... | placeholder | ... | john | ... | doe |
|---|---|---|---|---|---|---|---|---|---|
| $w$ | 1 | ... | 92 | ... | 5302 | ... | 13944 | ... | unk |
| $z_w$ | ∅ | | (birthd.,2,2) | | (occupation,1,1) | | (name,1,2) | | (name,2,1) (spouse,2,1) (children,2,1) |

**Figure 2:** Table features (right) for an example table (left); $\mathcal{W} \cup \mathcal{Q}$ is the set of all output words as defined in Section 3.3.

model. The table allows us to describe each word not only by its string (or index in the vocabulary) but also by a descriptor of its occurrence in the table. Let $\mathcal{F}$ define the set of all possible fields $f$. The occurrence of a word $w$ in the table is described by a set of (field, position) pairs.

$$z_w = \{(f_i, p_i)\}_{i=1}^m, \quad (3)$$

where $m$ is the number of occurrences of $w$. Each pair $(f, p)$ indicates that $w$ occurs in field $f$ at position $p$. In this scheme, most words are described by the empty set as they do not occur in the table. For example, the word *linguistics* in the table of Figure 1 is described as follows:

$$z_{\text{linguistics}} = \{(\text{fields}, 8); (\text{known for}, 4)\}, \quad (4)$$

assuming words are lower-cased and commas are treated as separate tokens.

Conditioning both on the field type and the position within the field allows the model to encode field-specific regularities, e.g., a number token in a date field is likely followed by a month token; knowing that the number is the first token in the date field makes this even more likely.

The (field, position) description scheme of the table does not allow to express that a token terminates a field which can be useful to capture field transitions. For biographies, the last token of the name field is often followed by an introduction of the birth date like '(' or 'was born'. We hence extend our descriptor to a triplet that includes the position of the token counted from the end of the field:

$$z_w = \{(f_i, p_i^+, p_i^-)\}_{i=1}^m, \quad (5)$$

where our example becomes:

$$z_{\text{linguistics}} = \{(\text{fields}, 8, 4); (\text{known for}, 4, 13)\}.$$

We extend Equation 2 to use the above information as additional conditioning context when generating a sentence $s$:

$$P(s|z) = \prod_{t=1}^T P(w_t|c_t, z_{c_t}), \quad (6)$$

where $z_{c_t} = z_{w_{t-(n-1)}}, \ldots, z_{w_{t-1}}$ are referred to as the local conditioning variables since they describe the local context (previous word) relations with the table.

**Global conditioning** refers to information from all tokens and fields of the table, regardless whether they appear in the previous generated words or not. The set of fields available in a table often impacts the structure of the generation. For biographies, the fields used to describe a politician are different from the ones for an actor or an athlete. We introduce global conditioning on the available fields $g_f$ as

$$P(s|z, g_f) = \prod_{t=1}^T P(w_t|c_t, z_{c_t}, g_f). \quad (7)$$

Similarly, global conditioning $g_w$ on the available

words occurring in the table is introduced:

$$P(s|z, g_f, g_w) = \prod_{t=1}^{T} P(w_t|c_t, z_{c_t}, g_f, g_w). \quad (8)$$

Tokens provide information complementary to fields. For example, it may be hard to distinguish a basketball player from a hockey player by looking only at the field names, e.g. teams, league, position, weight and height, etc. However the actual field tokens such as team names, league name, player's position can help the model to give a better prediction. Here, $g_f \in \{0,1\}^{\mathcal{F}}$ and $g_w \in \{0,1\}^{\mathcal{W}}$ are binary indicators over fixed field and word vocabularies.

Figure 2 illustrates the model with a schematic example. For predicting the next word $w_t$ after a given context $c_t$, the language model is conditioned on sets of triplets for each word occurring in the table $z_{c_t}$, along with all fields and words from this table.

## 3.3 Copy actions

So far we extended the model conditioning with features derived from the fact table. We now turn to using table information when scoring output words. In particular, sentences which express facts from a given table often copy words from the table. We therefore extend our model to also score special field tokens such as name_1 or name_2 which are subsequently added to the score of the corresponding words from the field value.

Our model reads a table and defines an output domain $\mathcal{W} \cup \mathcal{Q}$. $\mathcal{Q}$ defines all tokens in the table, which might include out of vocabulary words ($\notin \mathcal{W}$). For instance *Park-Rhodes* in Figure 1 is not in $\mathcal{W}$. However, *Park-Rhodes* will be included in $\mathcal{Q}$ as name_2 (since it is the second token of the name field) which allows our model to generate it. This mechanism is inspired by recent work on attention based word copying for neural machine translation (Luong et al., 2015) as well as delexicalization for neural dialog systems (Wen et al., 2015). It also builds upon older work such as class-based language models for dialog systems (Oh and Rudnicky, 2000).

## 4 A Neural Language Model Approach

A feed-forward neural language model (NLM) estimates $P(w_t|c_t)$ with a parametric function $\phi_\theta$ (Equation 1), where $\theta$ refers to all learnable parameters of the network. This function is a composition of simple differentiable functions or *layers*.

### 4.1 Mathematical notations and layers

We denote matrices as bold upper case letters ($\mathbf{X}$, $\mathbf{Y}$, $\mathbf{Z}$), and vectors as bold lower-case letters ($\mathbf{a}$, $\mathbf{b}$, $\mathbf{c}$). $\mathbf{A}_i$ represents the $i^{\text{th}}$ row of matrix $\mathbf{A}$. When $\mathbf{A}$ is a 3-d matrix, then $\mathbf{A}_{i,j}$ represents the vector of the $i^{\text{th}}$ first dimension and $j^{\text{th}}$ second dimension. Unless otherwise stated, vectors are assumed to be column vectors. We use $[\mathbf{v}_1; \mathbf{v}_2]$ to denote vector concatenation. Next, we introduce the notation for the different layers used in our approach.

**Embedding layer.** Given a parameter matrix $\mathbf{X} \in \mathbb{R}^{N \times d}$, the *embedding layer* is a lookup table that performs an array indexing operation:

$$\boldsymbol{\psi}_{\mathbf{X}}(x_i) = \mathbf{X}_i \in \mathbb{R}^d, \quad (9)$$

where $\mathbf{X}_i$ corresponds to the embedding of the element $x_i$ at row $i$. When $\mathbf{X}$ is a 3-d matrix, the lookup table takes two arguments:

$$\boldsymbol{\psi}_{\mathbf{X}}(x_i, x_j) = \mathbf{X}_{i,j} \in \mathbb{R}^d, \quad (10)$$

where $\mathbf{X}_{i,j}$ corresponds to the embedding of the pair $(x_i, x_j)$ at index $(i, j)$. The lookup table operation can be applied for a sequence of elements $s = x_1, \ldots, x_T$. A common approach is to concatenate all resulting embeddings:

$$\boldsymbol{\psi}_{\mathbf{X}}(s) = [\boldsymbol{\psi}_{\mathbf{X}}(x_1); \ldots; \boldsymbol{\psi}_{\mathbf{X}}(x_T)] \in \mathbb{R}^{T \times d}. \quad (11)$$

**Linear layer.** This layer applies a linear transformation to its inputs $\mathbf{x} \in \mathbb{R}^n$:

$$\boldsymbol{\gamma}_\theta(\mathbf{x}) = \mathbf{W}\mathbf{x} + \mathbf{b} \quad (12)$$

where $\theta = \{\mathbf{W}, \mathbf{b}\}$ are the trainable parameters with $\mathbf{W} \in \mathbb{R}^{m \times n}$ being the weight matrix, and $\mathbf{b} \in \mathbb{R}^m$ is the bias term.

**Softmax layer.** Given a context input $c_t$, the final layer outputs a score for each word $w_t \in \mathcal{W}$, $\boldsymbol{\phi}_\theta(c_t) \in \mathbb{R}^{|\mathcal{W}|}$. The probability distribution is obtained by applying the softmax activation function:

$$P(w_t = w|c_t) = \frac{\exp(\boldsymbol{\phi}_\theta(c_t, w))}{\sum_{i=1}^{|\mathcal{W}|} \exp(\boldsymbol{\phi}_\theta(c_t, w_i))} \quad (13)$$

### 4.2 Embeddings as inputs

A key aspect of neural language models is the use of word embeddings. Similar words tend to have similar embeddings and thus share latent features. The probability estimates of those models are smooth functions of these embeddings, and a small change in the features results in a small change in the probability estimates (Bengio et al., 2003). Therefore, neural language models can achieve better generalization for unseen n-grams. Next, we show how we map fact tables to continuous space in similar spirit.

**Word embeddings.** Formally, the embedding layer maps each context word index to a continuous $d$-dimensional vector. It relies on a parameter matrix $\mathbf{E} \in \mathbb{R}^{|\mathcal{W}| \times d}$ to convert the input $c_t$ into $n-1$ vectors of dimension $d$:

$$\boldsymbol{\psi}_{\mathbf{E}}(c_t) = \left[ \boldsymbol{\psi}_{\mathbf{E}}(w_{t-(n-1)}); \ldots; \boldsymbol{\psi}_{\mathbf{E}}(w_{t-1}) \right]. \quad (14)$$

$\mathbf{E}$ can be initialized randomly or with pre-trained word embeddings.

**Table embeddings.** As described in Section 3.2, the language model is conditioned on elements from the table. Embedding matrices are therefore defined to model both local and global conditioning information. For local conditioning, we denote the maximum length of a sequence of words as $l$. Each field $f_j \in \mathcal{F}$ is associated with $2 \times l$ vectors of $d$ dimensions, the first $l$ of those vectors embed all possible starting positions $1, \ldots, l$, and the remaining $l$ vectors embed ending positions. This results in two parameter matrices $\mathbf{Z} = \{\mathbf{Z}^+, \mathbf{Z}^-\} \in \mathbb{R}^{|\mathcal{F}| \times l \times d}$. For a given triplet $(f_j, p_i^+, p_i^-)$, $\boldsymbol{\psi}_{\mathbf{Z}^+}(f_j, p_i^+)$ and $\boldsymbol{\psi}_{\mathbf{Z}^-}(f_j, p_i^-)$ refer to the embedding vectors of the start and end position for field $f_j$, respectively.

Finally, global conditioning uses two parameter matrices $\mathbf{G}^f \in \mathbb{R}^{|\mathcal{F}| \times g}$ and $\mathbf{G}^w \in \mathbb{R}^{|\mathcal{W}| \times g}$. $\boldsymbol{\psi}_{\mathbf{G}^f}(f_j)$ maps a table field $f_j$ into a vector of dimension $g$, while $\boldsymbol{\psi}_{\mathbf{G}^w}(w_t)$ maps a word $w_t$ into a vector of the same dimension. In general, $\mathbf{G}^w$ shares its parameters with $\mathbf{E}$, provided $d = g$.

**Aggregating embeddings.** We represent each occurence of a word $w$ as a triplet (field, start, end) where we have embeddings for the start and end position as described above. Often times a particular word $w$ occurs multiple times in a table, e.g., 'linguistics' has two instances in Figure 1. In this case, we perform a component-wise max over the start embeddings of all instances of $w$ to obtain the best features across all occurrences of $w$. We do the same for end position embeddings:

$$\boldsymbol{\psi}_{\mathbf{Z}}(z_{w_t}) = \\ \left[ \max \left\{ \boldsymbol{\psi}_{\mathbf{Z}^+}(f_j, p_i^+), \forall (f_j, p_i^+, p_i^-) \in z_{w_t} \right\}; \right.\\ \left. \max \left\{ \boldsymbol{\psi}_{\mathbf{Z}^-}(f_j, p_i^-), \forall (f_j, p_i^+, p_i^-) \in z_{w_t} \right\} \right] \quad (15)$$

A special no-field embedding is assigned to $w_t$ when the word is not associated to any fields. An embedding $\boldsymbol{\psi}_{\mathbf{Z}}(z_{c_t})$ for encoding the local conditioning of the input $c_t$ is obtained by concatenation.

For global conditioning, we define $\mathcal{F}^q \subset \mathcal{F}$ as the set of all the fields in a given table $q$, and $\mathcal{Q}$ as the set of all words in $q$. We also perform max aggregation. This yields the vectors

$$\boldsymbol{\psi}_{\mathbf{G}^f}(g_f) = \max \left\{ \boldsymbol{\psi}_{\mathbf{G}^f}(f_j), \forall f_j \in \mathcal{F}^q \right\}, \quad (16)$$

and

$$\boldsymbol{\psi}_{\mathbf{G}^w}(g_w) = \max \left\{ \boldsymbol{\psi}_{\mathbf{G}^w}(w_t), \forall w_t \in \mathcal{Q} \right\}. \quad (17)$$

The final embedding which encodes the context input with conditioning is then the concatenation of these vectors:

$$\boldsymbol{\psi}_{\alpha_1}(c_t, z_{c_t}, g_f, g_w) = \left[ \boldsymbol{\psi}_{\mathbf{E}}(c_t); \boldsymbol{\psi}_{\mathbf{Z}}(z_{c_t}); \right.\\ \left. \boldsymbol{\psi}_{\mathbf{G}^f}(g_f); \boldsymbol{\psi}_{\mathbf{G}^w}(g_w) \right] \in \mathbb{R}^{d^1}, \quad (18)$$

with $\alpha_1 = \{\mathbf{E}, \mathbf{Z}^+, \mathbf{Z}^-, \mathbf{G}^f, \mathbf{G}^w\}$ and $d^1 = (n-1) \times (3 \times d) + (2 \times g)$. For simplification purpose, we define the context input $x = \{c_t, z_{c_t}, g_f, g_w\}$ in the following equations. This context embedding is mapped to a latent context representation using a linear operation followed by a hyperbolic tangent:

$$\mathbf{h}(x) = \tanh \left( \boldsymbol{\gamma}_{\alpha_2} \left( \boldsymbol{\psi}_{\alpha_1}(x) \right) \right) \in \mathbb{R}^{\text{nhu}}, \quad (19)$$

where $\alpha_2 = \{\mathbf{W}_2, \mathbf{b}_2\}$, with $\mathbf{W}_2 \in \mathbb{R}^{\text{nhu} \times d^1}$ and $\mathbf{b}_2 \in \mathbb{R}^{\text{nhu}}$.

### 4.3 In-vocabulary outputs

The hidden representation of the context then goes to another linear layer to produce a real value score for each word in the vocabulary:

$$\phi_\alpha^{\mathcal{W}}(x) = \boldsymbol{\gamma}_{\alpha_3} \left( \mathbf{h}(x) \right) \in \mathbb{R}^{|\mathcal{W}|}, \quad (20)$$

where $\alpha_3 = \{\mathbf{W}_3, \mathbf{b}_3\}$, with $\mathbf{W}_3 \in \mathbb{R}^{|\mathcal{W}| \times \text{nhu}}$ and $\mathbf{b}_3 \in \mathbb{R}^{|\mathcal{W}|}$, and $\alpha = \{\alpha_1, \alpha_2, \alpha_3\}$.

### 4.4 Mixing outputs for better copying

Section 3.3 explains that each word $w$ from the table is also associated with $z_w$, the set of fields in which it occurs, along with the position in that field. Similar to local conditioning, we represent each field and position pair $(f_j, p_i)$ with an embedding $\psi_\mathbf{F}(f_j, p_i)$, where $\mathbf{F} \in \mathbb{R}^{|\mathcal{F}| \times l \times d}$. These embeddings are then projected into the same space as the latent representation of context input $\mathbf{h}(x) \in \mathbb{R}^{\text{nhu}}$. Using the max operation over the embedding dimension, each word is finally embedded into a unique vector:

$$\mathbf{q}(w) = \max \{$$
$$\tanh\left(\boldsymbol{\gamma}_\beta\left(\boldsymbol{\psi}_\mathbf{F}(f_j, p_i)\right)\right), \forall (f_j, p_i) \in z_w\}, \quad (21)$$

where $\beta = \{\mathbf{W}_4, \mathbf{b}_4\}$ with $\mathbf{W}_4 \in \mathbb{R}^{\text{nhu} \times d}$, and $\mathbf{b}_4 \in \mathbb{R}^{\text{nhu}}$. A dot product with the context vector produces a score for each word $w$ in the table,

$$\phi_\beta^\mathcal{Q}(x, w) = \mathbf{h}(x) \cdot \mathbf{q}(w). \quad (22)$$

Each word $w \in \mathcal{W} \cup \mathcal{Q}$ receives a final score by summing the vocabulary score and the field score:

$$\phi_\theta(x, w) = \phi_\alpha^\mathcal{W}(x, w) + \phi_\beta^\mathcal{Q}(x, w), \quad (23)$$

with $\theta = \{\alpha, \beta\}$, and where $\phi_\beta^\mathcal{Q}(x, w) = 0$ when $w \notin \mathcal{Q}$. The softmax function then maps the scores to a distribution over $\mathcal{W} \cup \mathcal{Q}$,

$$\log P(w|x) = \phi_\theta(x, w) - \log \sum_{w' \in \mathcal{W} \cup \mathcal{Q}} \exp \phi_\theta(x, w').$$

### 4.5 Training

The neural language model is trained to minimize the negative log-likelihood of a training sentence $s$ with stochastic gradient descent (SGD; LeCun et al. 2012) :

$$L_\theta(s) = -\sum_{t=1}^{T} \log P(w_t | c_t, z_{c_t}, g_f, g_w). \quad (24)$$

## 5 Experiments

Our neural network model (Section 4) is designed to generate sentences from tables for large-scale problems, where a diverse set of sentence types need to be generated. Biographies are therefore a good framework to evaluate our model, with Wikipedia offering a large and diverse dataset.

### 5.1 Biography dataset

We introduce a new dataset for text generation, WIKIBIO, a corpus of 728,321 articles from English Wikipedia (Sep 2015). It comprises all biography articles listed by WikiProject Biography[1] which also have a table (infobox). We extract and tokenize the first sentence of each article with Stanford CoreNLP (Manning et al., 2014). All numbers are mapped to a special token, except for years which are mapped to different special token. Field values from tables are similarly tokenized. All tokens are lower-cased. Table 2 summarizes the dataset statistics: on average, the first sentence is twice as short as the table (26.1 vs 53.1 tokens), about a third of the sentence tokens (9.5) also occur in the table. The final corpus has been divided into three sub-parts to provide training (80%), validation (10%) and test sets (10%). The dataset is available for download[2].

### 5.2 Baseline

Our baseline is an interpolated Kneser-Ney (KN) language model and we use the KenLM toolkit to train 5-gram models without pruning (Heafield et al., 2013). We also learn a KN language model over templates. For that purpose, we replace the words occurring in both the table and the training sentences with a special token reflecting its table descriptor $z_w$ (Equation 3). The introduction section of the table in Figure 1 looks as follows under this scheme: "`name_1 name_2` ( `birthdate_1 birthdate_2 birthdate_3` – `deathdate_1 deathdate_2 deathdate_3` ) was an english linguist , `fields_3` pathologist , `fields_10` scientist , mathematician , mystic and mycologist ." During inference, the decoder is constrained to emit words from the regular vocabulary or special tokens occurring in the input table. When picking a special token we copy the corresponding word from the table.

### 5.3 Training setup

For our neural models, we train 11-gram language models ($n = 11$) with a learning rate set to 0.0025.

---
[1] https://en.wikipedia.org/wiki/Wikipedia:WikiProject_Biography
[2] https://github.com/DavidGrangier/wikipedia-biography-dataset

| Model | Perplexity | BLEU | ROUGE | NIST |
|---|---|---|---|---|
| KN | 10.51 | 2.21 | 0.38 | 0.93 |
| NLM | 9.40 ± 0.01 | 2.41 ± 0.33 | 0.52 ± 0.08 | 1.27 ± 0.26 |
| + Local (field, start, end) | 8.61 ± 0.01 | 4.17 ± 0.54 | 1.48 ± 0.23 | 1.41 ± 0.11 |
| Template KN | 7.46★ | 19.8 | 10.7 | 5.19 |
| Table NLM w/ Local (field, start) | 4.60 ± 0.01† | 26.0 ± 0.39 | 19.2 ± 0.23 | 6.08 ± 0.08 |
| + Local (field, start, end) | 4.60 ± 0.01† | 26.6 ± 0.42 | 19.7 ± 0.25 | 6.20 ± 0.09 |
| + Global (field) | **4.30 ± 0.01†** | 33.4 ± 0.18 | 23.9 ± 0.12 | 7.52 ± 0.03 |
| + Global (field & word) | 4.40 ± 0.02† | **34.7 ± 0.36** | **25.8 ± 0.36** | **7.98 ± 0.07** |

Table 1: BLEU, ROUGE, NIST and perplexity without copy actions (first three rows) and with copy actions (last five rows). For neural models we report "mean ± standard deviation" for five training runs with different initialization. Decoding beam width is 5. Perplexities marked with ★ and † are not directly comparable as the output vocabularies differ slightly.

| | Mean | Percentile | |
|---|---|---|---|
| | | 5% | 95% |
| # tokens per sentence | 26.1 | 13 | 46 |
| # tokens per table | 53.1 | 20 | 108 |
| # table tokens per sent. | 9.5 | 3 | 19 |
| # fields per table | 19.7 | 9 | 36 |

Table 2: Dataset statistics

| Parameter | | Value | |
|---|---|---|---|
| # word types | $\|\mathcal{W}\|$ | = | 20,000 |
| # field types | $\|\mathcal{F}\|$ | = | 1,740 |
| Max. # tokens in a field | $l$ | = | 10 |
| word/field embedding size | $d$ | = | 64 |
| global embedding size | $g$ | = | 128 |
| # hidden units | nhu | = | 256 |

Table 3: Model Hyperparameters

Table 3 describes the other hyper-parameters. We include all fields occurring at least 100 times in the training data in $\mathcal{F}$, the set of fields. We include the 20,000 most frequent words in the vocabulary. The other hyperparameters are set through validation, maximizing BLEU over a validation subset of 1,000 sentences. Similarly, early stopping is applied: training ends when BLEU stops improving on the same validation subset. One should note that the maximum number of tokens in a field $l = 10$ means that we encode only 10 positions: for longer field values the final tokens are not dropped but their position is capped to 10. We initialize the word embeddings $W$ from Hellinger PCA computed over the set of training biographies. This representation has shown to be helpful for various applications (Lebret and Collobert, 2014).

### 5.4 Evaluation metrics

We use different metrics to evaluate our models. Performance is first evaluated in terms of perplexity which is the standard metric for language modeling. Generation quality is assessed automatically with BLEU-4, ROUGE-4 (F-measure) and NIST-4[3] (Belz and Reiter, 2006).

## 6 Results

This section describes our results and discusses the impact of the different conditioning variables.

### 6.1 The more, the better

The results (Table 1) show that more conditioning information helps to improve the performance of our models. The generation metrics BLEU, ROUGE and NIST all gives the same performance ordering over models. We first discuss models without copy actions (the first three results) and then discuss models with copy actions (the remaining results). Note that the factorization of our models results in three different output domains which makes perplexity comparisons less straightforward: models without copy actions operate over a fixed vocabulary. Template KN adds a fixed set of field/position pairs to this vocabulary while Table NLM models a variable set $\mathcal{Q}$ depending on the input table, see Section 3.3.

**Without copy actions.** In terms of perplexity the (i) neural language model (NLM) is slightly better than an interpolated KN language model, and (ii) adding local conditioning on the field start and end position further improves accuracy. Generation metrics are generally very low but there is a clear improvement when using local conditioning since it allows to learn transitions between fields by linking previous predictions to the table unlike KN or plain NLM.

---
[3] We rely on standard software, NIST mteval-v13a.pl (for NIST, BLEU), and MSR rouge-1.5.5 (for ROUGE).

**With copy actions.** For experiments with copy actions we use the full local conditioning (Equation 4) in the neural language models. BLEU, ROUGE and NIST all improves when moving from Template KN to Table NLM and more features successively improve accuracy. Global conditioning on the fields improves the model by over 7 BLEU and adding words gives an additional 1.3 BLEU. This is a total improvement of nearly 15 BLEU over the Template Kneser-Ney baseline. Similar observations are made for ROUGE +15 and NIST +2.8.

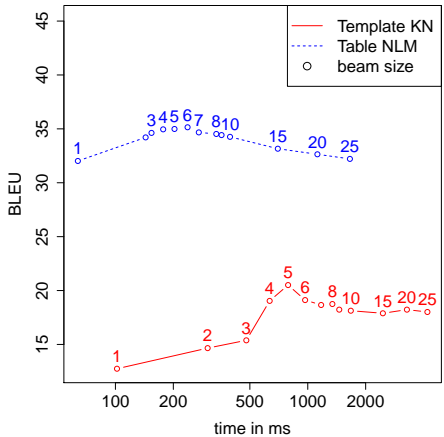

**Figure 3:** Comparison between our best model (Table NLM) and the baseline (Template KN) for different beam sizes. The x-axis is the average timing (in milliseconds) for generating one sentence. The y-axis is the BLEU score. All results are measured on a subset of 1,000 samples of the validation set.

### 6.2 Attention mechanism

Our model implements attention over input table fields. For each word $w$ in the table, Equation (23) takes the language model score $\phi_{c_t}^{\mathcal{W}}$ and adds a bias $\phi_{c_t}^{\mathcal{Q}}$. The bias is the dot-product between a representation of the table field in which $w$ occurs and a representation of the context, Equation (22) that summarizes the previously generated fields and words.

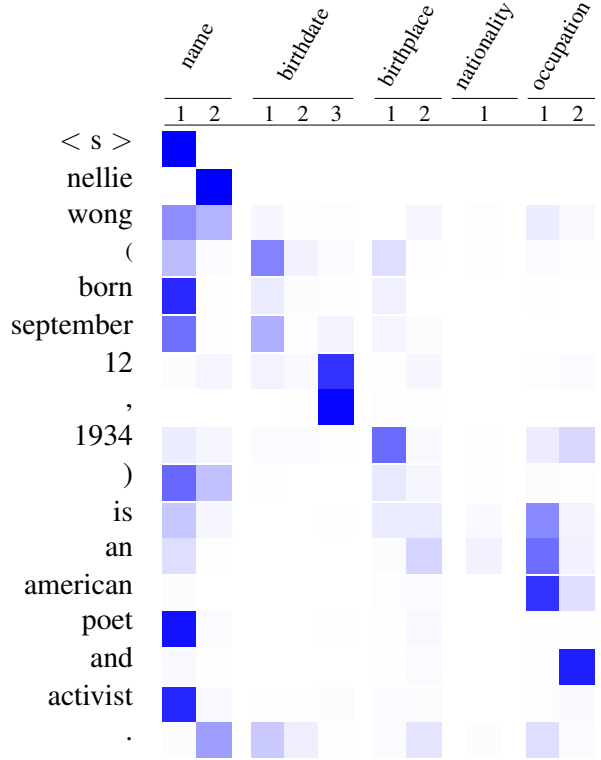

**Figure 4:** Visualization of attention scores for Nellie Wong's Wikipedia infobox. Each row represents the probability distribution over (field, position) pairs given the previous words (i.e. the words heading the preceding rows as well as the current row). Darker colors depict higher probabilities.

Figure 4 shows that this mechanism adds a large bias to continue a field if it has not generated all tokens from the table, e.g., it emits the word occurring in name_2 after generating name_1. It also nicely handles transitions between field types, e.g., the model adds a large bias to the words occurring in the *occupation* field after emitting the *birthdate*.

### 6.3 Sentence decoding

We use a standard beam search to explore a larger set of sentences compared to simple greedy search. This allows us to explore $K$ times more paths which comes at a linear increase in the number of forward computation steps for our language model. We compare various beam settings for the baseline Template KN and our Table NLM (Figure 3). The best validation BLEU can be obtained with a beam size of $K = 5$. Our model is also several times faster than the baseline, requiring only about 200 ms per sentence with $K = 5$. Beam search generates many n-gram lookups for Kneser-Ney which requires many random memory accesses; while neural models perform scoring through matrix-matrix products, an operation which is more local and can be performed in a block parallel manner where modern graphic processors shine (Kindratenko, 2014).

| Model | Generated Sentence |
|---|---|
| Reference | frederick parker-rhodes (21 march 1914 – 21 november 1987) was an english linguist, plant pathologist, computer scientist, mathematician, mystic, and mycologist. |
| Baseline (Template KN) | frederick parker-rhodes ( born november 21 , 1914 – march 2 , 1987 ) was an english cricketer . |
| Table NLM +Local (field, start) | frederick parker-rhodes ( 21 november 1914 – 2 march 1987 ) was an australian rules footballer who played with carlton in the victorian football league ( vfl ) during the XXXXs and XXXXs . |
| + Global (field) | frederick parker-rhodes ( 21 november 1914 – 2 march 1987 ) was an english mycology and plant pathology , mathematics at the university of uk . |
| + Global (field, word) | frederick parker-rhodes ( 21 november 1914 – 2 march 1987 ) was a british computer scientist , best known for his contributions to computational linguistics . |

Table 4: First sentence from the current Wikipedia article about Frederick Parker-Rhodes and the sentences generated from the three versions of our table-conditioned neural language model (Table NLM) using the Wikipedia infobox seen in Figure 1.

### 6.4 Qualitative analysis

Table 4 shows generations for different variants of our model based on the Wikipedia table in Figure 1. First of all, comparing the reference to the fact table reveals that our training data is not perfect. The birth month mentioned in the fact table and the first sentence of the Wikipedia article are different; this may have been introduced by one contributor editing the article and not keeping the information consistent.

All three versions of our model correctly generate the beginning of the sentence by copying the name, the birth date and the death date from the table. The model correctly uses the past tense since the death date in the table indicates that the person has passed away. Frederick Parker-Rhodes was a scientist, but this occupation is not directly mentioned in the table. The model without global conditioning can therefore not predict the right occupation, and it continues the generation with the most common occupation (in Wikipedia) for a person who has died. In contrast, the global conditioning over the fields helps the model to understand that this person was indeed a scientist. However, it is only with the global conditioning on the words that the model can infer the correct occupation, i.e., *computer scientist*.

## 7 Conclusions

We have shown that our model can generate fluent descriptions of arbitrary people based on structured data. Local and global conditioning improves our model by a large margin and we outperform a Kneser-Ney language model by nearly 15 BLEU. Our task uses an order of magnitude more data than previous work and has a vocabulary that is three orders of magnitude larger.

In this paper, we have only focused on generating the first sentence and we will tackle the generation of longer biographies in future work. Also, the encoding of field values can be improved. Currently, we only attach the field type and token position to each word type and perform a max-pooling for local conditioning. One could leverage a richer representation by learning an encoder conditioned on the field type, e.g. a recurrent encoder or a convolutional encoder with different pooling strategies.

Furthermore, the current training loss function does not explicitly penalize the model for generating incorrect facts, e.g. predicting an incorrect nationality or occupation is currently not considered worse than choosing an incorrect determiner. A loss function that could assess factual accuracy would certainly improve sentence generation by avoiding such mistakes. Also it will be important to define a strategy for evaluating the factual accuracy of a generation, beyond BLEU, ROUGE or NIST.